\documentclass{article}
\ifdefined\pdfminorversion
\fi
\usepackage{iclr2027_conference,times}
\usepackage{amsmath,amssymb}
\usepackage{booktabs}
\usepackage{graphicx}
\usepackage{adjustbox}
\usepackage{microtype}
\usepackage{enumitem}
\usepackage{hyperref}
\usepackage{url}

\title{Compress What You See, Not What You Say: Anchored Context Distillation for Latent-Observation Software Engineering Agents}

\author{Zhensheng Zou \quad Guoqing Wang \quad Dan Hao \\ Peking University}

\iclrfinalcopy
\hypersetup{
  hidelinks,
  pdftitle={Compress What You See, Not What You Say: Anchored Context Distillation for Latent-Observation Software Engineering Agents},
  pdfauthor={Zhensheng Zou, Guoqing Wang, Dan Hao}
}

\begin{document}

\maketitle
\pagestyle{plain}

\begin{abstract}

Tool observations dominate the context of software-engineering agents, making long interaction histories costly to maintain. Existing context compression methods can discard information needed by later actions, while adapting agents to soft-token representations can compromise their original behavior. To reduce context while preserving action-critical information and agent behavior, we combine \emph{Latent Observations, Hard Actions} (LOHA), a context layout that separates compressed history from text needed for exact reference, with \emph{Anchored Context Distillation} (ACD), a training method that enables latent reading while constraining behavioral drift.
LOHA compresses older tool observations into soft tokens while retaining the agent's own turns and the last $K$ observations in text, providing compact access to historical information and exact access to recent content. To enable the agent to use this representation, ACD distills the base model's full-text predictions into the latent view while anchoring its behavior on plain-text inputs to the same base model.
On SWE-bench Verified, $K{=}3$ reduces context per call by 43\% for Qwen3-4B and 57\% for SWE-Master-4B-RL, with resolve rates of 12.1\% and 21.8\% versus 14.5\% and 27.5\% for their uncompressed bases. A single-run recency sweep reaches 14.4\% and 23.0\% at $K{=}8$, with larger windows generally favoring task performance over compression. Under a 32K-token limit, Qwen3 with $K{=}3$ resolves 21.1\% of a 199-instance subset versus 11.1\% for the same adapted agent using full text. In concurrent single-GPU serving, it achieves $1.9\times$ that full-text agent's instance throughput.
\end{abstract}

\section{Introduction}
\label{sec:introduction}

Large language models are increasingly used as software-engineering agents. To address a GitHub issue, an agent uses tools to inspect the repository, modify source code, and test its changes, refining the patch based on the results~\citep{jimenez2024swebench,yang2024sweagent,xia2024agentless,wang2024openhands}. These interactions produce tool observations, such as file contents, search results, stack traces, and test logs, that accumulate in the agent's context. In CoderForge trajectories~\citep{together2026coderforge}, tool observations account for 69\% of trajectory tokens on average.

Existing approaches reduce context through observation masking~\citep{lindenbauer2025complexity}, history summarization~\citep{packer2023memgpt,kang2025acon}, sub-task folding~\citep{sun2025contextfolding,ye2025agentfold}, or selective pruning of tool outputs~\citep{wang2026swepruner,wang2026sweprunerpro,ren2026taco,chen2026coact}. These methods shorten interaction histories, but deciding what to discard is difficult when later actions may depend on details in earlier observations. Coding agents also require more than a semantic summary: a \texttt{str\_replace} edit, for example, requires an exact match to the source text. Information omitted during compression may therefore have to be retrieved again before the agent can act. Source-code compressors~\citep{shi2025longcodezip,wang2024slimcode} exploit program structure to guide compression, but are less suited to the search results, stack traces, and test logs that also occupy a substantial portion of an agent's context.

Soft-token compression~\citep{mu2023gist,ge2024icae,li2024500xcompressor,li2026lclm} offers an alternative by encoding text into a shorter sequence of continuous embeddings that the decoder reads directly. For coding agents, however, a useful compressed representation must support both understanding past observations and reproducing specific details for tool calls. These requirements are distinct: retaining enough information to reason about a file does not necessarily preserve the exact strings needed to edit it. Adopting soft-token representations also requires the decoder to learn a new input representation while retaining its existing ability to reason, invoke tools, and modify code. This raises two closely related questions: how should an agent combine compressed history with access to exact text, and how can it learn to use that history without compromising its ability to act?

\begin{figure}[t]
\centering
\includegraphics[width=0.98\linewidth]{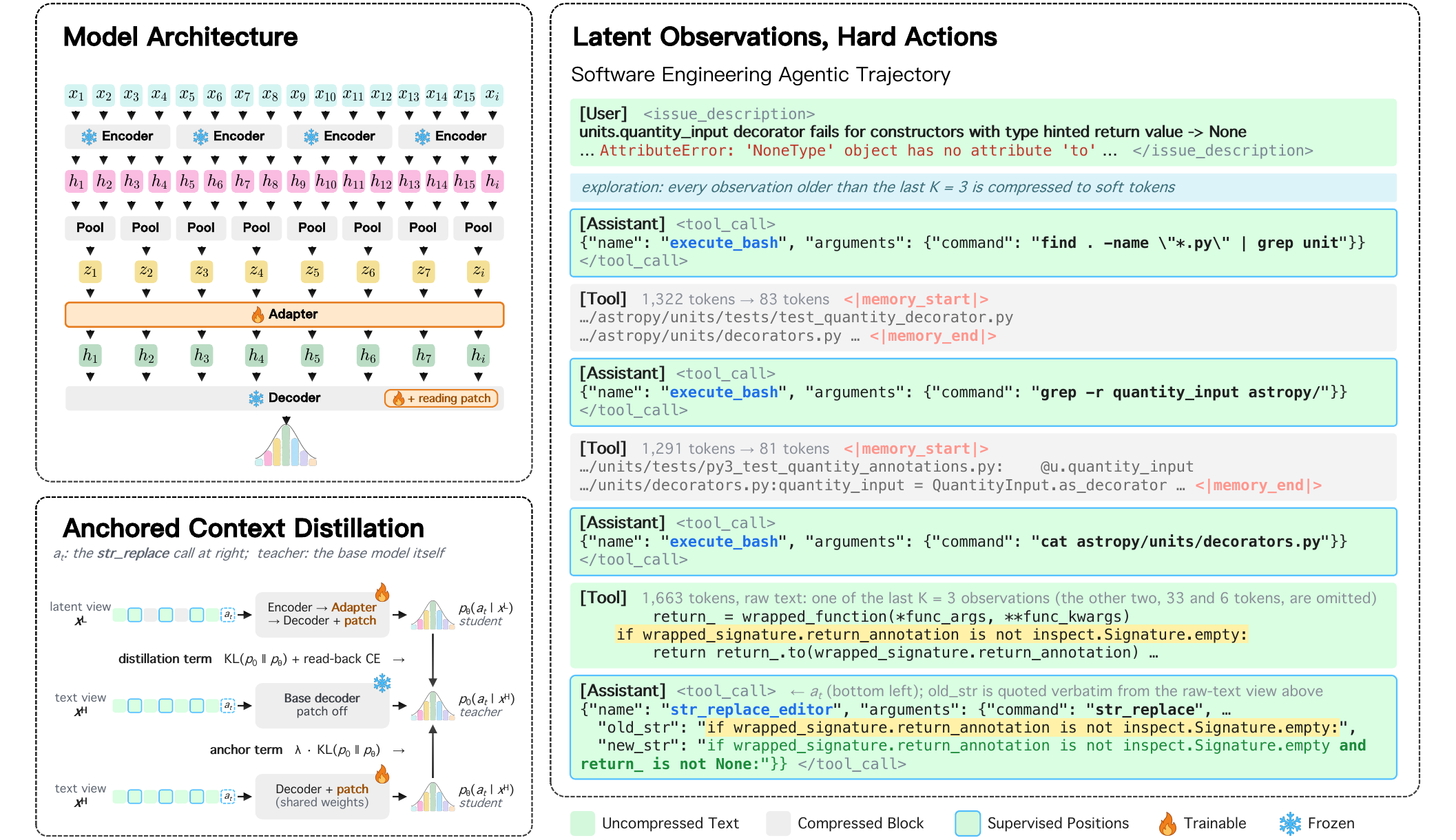}
\caption{LOHA retains recent text and compresses older observations; ACD adds latent reading with a full-text behavioral anchor. Only the adapter, reading patch, and normalization gains are trained.}
\label{fig:overview}
\end{figure}

We address these questions with a framework that combines context compression with behavior-preserving adaptation (Figure~\ref{fig:overview}). Its two components, \emph{Latent Observations, Hard Actions} (LOHA) and \emph{Anchored Context Distillation} (ACD), determine how the agent's history is represented and how the agent learns to use it. LOHA follows a simple principle: \textbf{compress what the agent sees, not what it says}. It represents older tool observations as soft tokens while keeping the agent's own turns, the system prompt, and the task description in ordinary text. To support actions that require exact strings, it also retains the $K$ most recent observations in raw text. This layout provides a compact representation of historical observations while preserving direct access to recent content for precise tool calls. ACD then enables the agent to read the compressed observations while limiting changes to its existing behavior. It combines two complementary objectives: a distillation term aligns predictions from the latent view with those of the frozen base model given the full text, while an anchor term constrains the adapted model's predictions on full-text inputs to remain close to the same base model. Together, LOHA and ACD address the representation and adaptation challenges of using compressed observations in coding agents.

We evaluate LOHA + ACD on SWE-bench Verified with Qwen3-4B-Instruct and its RL-fine-tuned derivative, SWE-Master-4B-RL. At their full context windows, $K{=}3$ reduces context per call by \textbf{43\%} and \textbf{57\%}, with resolve rates of \textbf{12.1\%} and \textbf{21.8\%} versus \textbf{14.5\%} and \textbf{27.5\%} for their uncompressed bases. In the single-run recency sweep, $K{=}8$ resolves 14.4\% and 23.0\%, versus 13.6\% and 22.2\% at $K{=}3$, with more text retained and higher compute cost (Table~\ref{tab:sweeps-main}a). On SWE-Master, $K{=}3$ also outperforms masking (17.6\%). Under a 32K-token limit, Qwen3 with $K{=}3$ resolves \textbf{21.1\%} of a 199-instance subset versus \textbf{11.1\%} for the adapted full-text agent. Concurrent single-GPU serving achieves \textbf{$1.9\times$} that agent's instance throughput.
Our contributions are:
\begin{itemize}[leftmargin=1.2em,itemsep=1pt,topsep=2pt]
  \item \textbf{A context layout for latent-observation coding agents.}
  LOHA combines soft-token representations of older tool observations with uncompressed agent turns and a recent-observation window, providing compact historical context alongside exact text for actions involving recent content.

  \item \textbf{Anchored Context Distillation for behavior-preserving adaptation.}
  ACD combines cross-view distillation with a self-anchoring objective to teach latent reading while limiting behavioral drift. Our experiments show that better read-back performance alone does not ensure better agent behavior and identify self-anchoring as an effective way to balance the two.

  \item \textbf{An empirical evaluation of the performance--efficiency trade-off.}
  We compare task success and context usage on both an instruction-tuned agent and an RL-fine-tuned agent, and examine context limits and serving throughput. The results characterize the performance cost of compression and the settings in which it provides practical benefits.
\end{itemize}

\section{Approach}
\label{sec:approach}

Our framework combines two components: \emph{Latent Observations, Hard Actions} (LOHA), which organizes the agent's context into compressed observations and uncompressed text, and \emph{Anchored Context Distillation} (ACD), which trains the agent to use this representation while limiting changes to its existing behavior. We first define the context layout, then describe the student and its objectives, and finally explain training, inference, and cost accounting.

\subsection{Problem Setup}
\label{sec:setup}

An issue-resolution trajectory consists of a system prompt $s$ containing tool schemas, an issue description $u$, assistant turns $a_t$, and tool observations $o_t$: $\tau=(s,u,a_1,o_1,\ldots,a_T)$. Assistant turns contain reasoning and tool calls; observations contain information returned by the environment, such as file contents, search results, and test logs.

We use an LCLM-based compression pipeline~\citep{li2026lclm}, denoted by $\mathcal{E}$, comprising an encoder and an adapter that maps its output to the decoder's input space. At the default $16\times$ ratio, $n$ text tokens become approximately $n/16$ continuous embeddings (\emph{soft tokens}). The decoder processes these alongside ordinary text-token embeddings (\emph{hard tokens}).

\subsection{Latent Observations, Hard Actions}
\label{sec:loha}

LOHA compresses older tool observations while retaining the system prompt, issue description, and all assistant turns in text. A recent-observation window also remains uncompressed, providing exact text for actions involving recently accessed content.

\textbf{Rendering.}
Each observation remains in its original position within the decoder's native tool-response envelope. For an observation selected for compression, its body is wrapped in memory markers, encoded in 1{,}024-token windows, and replaced by soft tokens. The surrounding envelope remains in hard tokens, preserving its association with the tool call; no additional header is introduced. Unlike pruning, this represents the observation through an encoder rather than selecting text spans, though compression does not guarantee verbatim reconstruction.

\textbf{The recency policy (Hard-Last-$K$).}
An edit often follows a file view, and a test command may quote an identifier from a recent failure report. LOHA therefore uses recency to retain raw text, rather than predicting which observation the next action will quote. Given $m$ observations, the body of $o_i$ is represented as
\[
r_K(o_i\mid m)=
\begin{cases}
o_i, & i>m-K,\\
\mathcal{E}(o_i), & \text{otherwise}.
\end{cases}
\]
For $K>0$, each new observation enters as text and becomes eligible for compression when it leaves the window. Older observations remain available through their latent representations; exact content can be retrieved again using the existing tools. The window size is selected at inference without retraining: increasing $K$ retains more exact text but reduces compression for a fixed history. We use $K=3$ for the main comparisons and evaluate $K\in\{0,1,2,3,4,8,\infty\}$, spanning fully latent to full-text observations (Section~\ref{sec:eval:ablations}).

\subsection{Anchored Context Distillation}
\label{sec:acd}

ACD adapts a base agent to read latent observations while regularizing its predictions on ordinary text. We apply the same recipe to Qwen3-4B-Instruct-2507~\citep{qwen3} and SWE-Master-4B-RL, each serving as the base decoder for its own student and as its frozen teacher. Rather than adopting the continually pre-trained LCLM decoder directly, we use its weight difference from the base decoder to initialize a low-rank reading patch.

The training objective covers two input views. On the latent view, the student learns to match the teacher's predictions from the corresponding full-text context. On the full-text view, the student is regularized toward the teacher's original predictions. These objectives address latent reading and behavioral preservation within the same adapted model.

\begin{figure}[t]
\centering
\includegraphics[width=\textwidth]{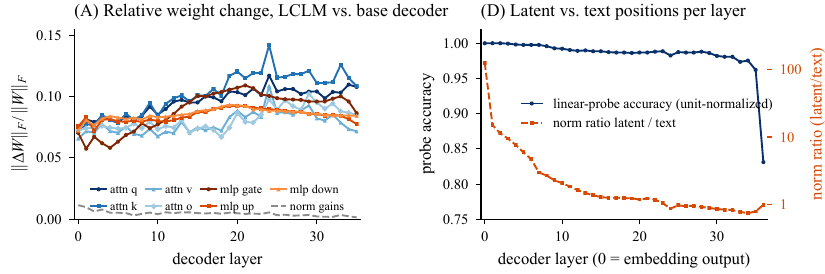}
\caption{(A) Projection-weight and normalization-gain changes from Qwen3 to LCLM. (D) Unit-normalized latent/text probe accuracy and mean norm ratio. Full measurements: Appendix~\ref{app:mechanism}.}
\label{fig:mechanism}
\end{figure}

\textbf{Design motivation.}
Soft tokens introduce a new input representation, so the decoder must learn to use them while retaining its existing tool-use policy. The LCLM decoder supplies a trained reading initialization, but adopting it directly would also replace the base agent's decoder. A low-rank patch provides a limited set of parameters through which to transfer and adapt that capability.

Both latent and full-text inputs use the adapted decoder, so a reading objective can also change its predictions on ordinary text. ACD therefore adds a full-text anchor that directly regularizes those predictions. Figure~\ref{fig:mechanism} summarizes diagnostics of the decoder weights and latent inputs; quantitative measurements and initialization probes are reported in Appendix~\ref{app:mechanism}.

\subsubsection{The Student}
\label{sec:acd:student}

The student contains a frozen LCLM encoder, a trainable MLP adapter, and the base decoder augmented with a low-rank reading patch. Decoder normalization gains are also trainable and initialized from the base model; its remaining parameters are frozen.

For each projection $W_0$, we define $\Delta W=W_{\mathrm{LCLM}}-W_0$ and parameterize the adapted weight as $W_\theta=W_0+AB$, with $\operatorname{rank}(AB)\leq\rho$.
The factors $A$ and $B$ are initialized so that their product equals the rank-$\rho$ truncated singular value decomposition of $\Delta W$. The patch covers all 252 decoder projections. At the default rank $\rho=64$, it has 132M parameters, and the adapter has 9.2M.

We denote the trainable patch, adapter, and gains by $\theta$, the student by $p_\theta$, and the unmodified teacher by $p_0$. Teacher predictions are computed before student training and cached. The student uses the same adapted decoder in both views; the compression encoder and adapter are used only when constructing latent inputs.

\subsubsection{Two Views, Two Terms}
\label{sec:acd:objective}

Each trajectory is rendered as a \emph{full-text view} $x^{\mathrm{H}}$, with all observations in hard tokens, and a \emph{latent view} $x^{\mathrm{L}}$, with all observation bodies compressed. Let $\mathcal{J}$ index corresponding assistant tokens and $c_j^{\mathrm{H}}$, $c_j^{\mathrm{L}}$ their preceding contexts. Predictions are aligned by assistant-token identity, since compression changes absolute positions.

We also append read-back questions to latent examples. Let $\mathcal{R}$ index their answer tokens, with target $y_r$ and preceding context $c_r^{\mathrm{L}}$. These positions directly supervise recovery of content from compressed observations.

The objective is $\mathcal{L}=\mathcal{L}_{\mathrm{distill}}+\lambda\mathcal{L}_{\mathrm{anchor}}$, where
\begin{align}
\mathcal{L}_{\mathrm{distill}} &= \frac{1}{N_{\mathrm{distill}}}\Bigg[\sum_{j\in\mathcal{J}} D_{\mathrm{KL}}\!\left(p_0(\cdot\mid c_j^{\mathrm{H}})\,\middle\|\,p_\theta(\cdot\mid c_j^{\mathrm{L}})\right)-\sum_{r\in\mathcal{R}}\log p_\theta(y_r\mid c_r^{\mathrm{L}})\Bigg],
\label{eq:acd-distill}\\
\mathcal{L}_{\mathrm{anchor}}
&=
\frac{1}{N_{\mathrm{anchor}}}
\sum_{j\in\mathcal{J}}
D_{\mathrm{KL}}\!\left(
p_0(\cdot\mid c_j^{\mathrm{H}})
\,\middle\|\,
p_\theta(\cdot\mid c_j^{\mathrm{H}})
\right).
\label{eq:acd-anchor}
\end{align}
Here $N_{\mathrm{distill}}=|\mathcal{J}|+|\mathcal{R}|$ and $N_{\mathrm{anchor}}=|\mathcal{J}|$ count supervised positions within a batch. We use forward KL by default; alternative directions are evaluated in Appendix~\ref{app:results:rq2}.

\textbf{Learning from the latent view.}
The distillation term encourages the student to make predictions from compressed observations that are consistent with the base model's predictions from the original text. Matching the teacher's distribution provides supervision beyond the single action recorded in the trajectory.

The read-back component adds explicit supervision for extracting literal content from compressed observations. Its answers are substrings of the source observations and account for approximately 3\% of supervised positions. The construction and filtering of these questions are described in Appendix~\ref{app:method}.

\textbf{Anchoring on the full-text view.}
The anchor term regularizes the adapted model on inputs that contain no latent observations. Cross-view distillation alone does not directly constrain predictions on these full-text inputs, even though both views use the same adapted decoder.

Because the full-text view bypasses the compression pipeline, the anchor term updates the reading patch and normalization gains but has no gradient with respect to the compression adapter. The latent-view loss updates all three components. Separately normalizing the two terms makes their relative weight explicit through $\lambda$, rather than letting it depend on the numbers of supervised tokens in the two views. We use $\lambda=1$ in all reported runs.

\textbf{Choice of teacher.}
Both terms use the original base model as their teacher, but for different purposes: it provides the full-text prediction target for latent reading and the reference distribution for behavioral preservation. Using an external model for the anchor would instead encourage the student to adopt that model's full-text predictions. Self-anchoring directly expresses our objective of retaining the starting agent's behavior while adding latent-reading capability. The distillation teacher need not be identical to the anchor; Appendix~\ref{app:results:rq2} examines their roles separately.

\textbf{Estimating the divergences.}
Since the teacher is frozen and the training trajectories are fixed, its predictions can be computed once. At each supervised assistant position, we cache the teacher's top-$k_{\mathrm{logits}}$ token log-probabilities, with $k_{\mathrm{logits}}=64$, together with the remaining probability mass.
During training, we evaluate each divergence over the cached tokens and one additional bucket containing the rest of the vocabulary. This coarsened divergence matches the probability mass assigned to the tail but does not constrain its internal distribution. Before numerical safeguards, it is a lower bound on the full-vocabulary KL in either direction. Implementation details and approximation measurements are in Appendices~\ref{app:method} and~\ref{app:mechanism:kl}.

\subsection{Training and Inference}
\label{sec:training-inference}

\textbf{Training.}
We select the shortest successful CoderForge trajectory per task~\citep{together2026coderforge}, yielding 35{,}140 trajectories; ablations use the first 3{,}514. Trajectories are rendered in both views and packed into 204{,}800-token sequences.
We train for one epoch with AdamW, using a learning rate of $10^{-4}$ for the reading patch and normalization gains and $5\times10^{-5}$ for the adapter. Training is offline: predictions are evaluated on recorded trajectory prefixes rather than on interactions generated by the student. Cached targets avoid a teacher forward pass during optimization, while gradients propagate through the student decoder to its trainable components.

A plain-text tool-calling probe checks editing behavior and tool-call validity beyond the training losses. All reported arms are evaluated regardless of probe outcomes. Further optimization and probe details are in Appendix~\ref{app:method}.

\textbf{Inference.}
The adapted agent uses the scaffold's native function-calling interface and ordinary tools. Before each model call, LOHA renders the history with the selected $K$; observations shorter than 128 characters remain in hard tokens regardless of age. Training thus uses the fully latent and full-text endpoints, whereas mixed-window inference combines older latent observations with recent raw text.

\textbf{Cost accounting.}
We distinguish representation length from the cost of an executed trajectory. Encoded observation bodies use approximately one soft token per 16 source tokens. The context-level reduction is smaller because envelopes, short observations, prompts, assistant turns, and the recent window remain in text.

For a fixed trajectory, we compare the compressed representation with the full-text rendering of the same history. Separately, we measure cumulative decoder input over actual rollouts, whose lengths and tool calls may differ across conditions.

Advancing the recent window changes an existing prefix and invalidates its cache from the first modified observation. The current server also re-encodes observations on each call. We therefore report encoder work, prefix reuse, and serving throughput separately from context length (Section~\ref{sec:eval:rq4}); metric definitions are in Appendix~\ref{app:protocol}.

\section{Experiments}
\label{sec:evaluation}

We evaluate task performance and context cost on two base agents, Qwen3-4B-Instruct-2507 and SWE-Master-4B-RL, its RL-fine-tuned derivative. Both use the same LOHA + ACD recipe, with each agent anchored to its own unmodified model. We sweep the recency window for both families; further Qwen3 experiments examine context limits, anchor identity, and serving efficiency.

\textbf{Setup.}
We evaluate both model families on 499 instances of SWE-bench Verified~\citep{jimenez2024swebench} using OpenHands~\citep{wang2024openhands} 0.62.0. Qwen3 uses a 262K-token context window and a 200-iteration cap; SWE-Master uses 131K tokens and 100 iterations. Conditions within each family share the scaffold and decoding settings, with one attempt per instance per seed. The uncompressed base, observation masking, and the two ACD conditions use two or three seeds; pruning baselines use one run. Full protocols are in Appendix~\ref{app:protocol}.

\textbf{Comparisons.}
The main comparisons use LOHA + ACD at $K{=}3$, the uncompressed base agent, and the same adapted model using full text ($K{=}\infty$). The latter separates adaptation from the effect of context compression. Baselines are observation masking~\citep{lindenbauer2025complexity}, SWE-Pruner~\citep{wang2026swepruner}, Self-Prune, and LongCodeZip~\citep{shi2025longcodezip}, all using the same recent-observation window and compression threshold.

\phantomsection\textbf{Metrics.}
\label{sec:metrics}
We report resolve rate, context tokens per call, decoder input tokens per trajectory, and serving throughput. The whole-trajectory compression ratio compares the plaintext equivalent of an agent's own contexts with the tokens it actually reads. Encoder work, prefix-cache reuse, and estimated compute are reported separately; metric definitions and paired-test results are in Appendices~\ref{app:protocol} and~\ref{app:results}.

\subsection{Full-Window Results}
\label{sec:eval:rq1}

At $K{=}3$, LOHA reduces context per call by 43\% for Qwen3 and 57\% for SWE-Master, with a performance cost in both families (Tables~\ref{tab:rq1} and~\ref{tab:transfer}). The recency sweep includes $K{=}8$ as a larger-window alternative (Table~\ref{tab:sweeps-main}a).

\begin{table*}[t]
\centering
\caption{Qwen3-4B-Instruct-2507 (262K context, 200 iterations). Resolved: out of 499, mean $\pm$ standard deviation; pruners use single runs. Cost definitions: Appendix~\ref{app:protocol}.}
\label{tab:rq1}
\scriptsize
\setlength{\tabcolsep}{2.5pt}
\begin{adjustbox}{max width=\textwidth}\begin{tabular}{@{}l r r r r r r r r r@{}}
\toprule
\textbf{Condition} & \textbf{resolved} & \textbf{calls} & \textbf{in/call} & \textbf{dec in} & \textbf{cached} & \textbf{enc/aux tok} & \textbf{ratio} & \textbf{dec PF} & \textbf{enc/aux PF} \\
\midrule
Uncompressed anchor & 72.3 $\pm$ 3.1 (14.5\%) & 37.9 & 43.6K & 1.65M & 96.9\% & --- & $1.00\times$ & 1.82 & --- \\
\midrule
\multicolumn{10}{@{}l}{\textit{LOHA + ACD (ours)}} \\
\hspace{0.8em}plain text ($K{=}\infty$) & 66.0 $\pm$ 2.8 (13.2\%) & 48.0 & 50.9K & 2.45M & 95.2\% & --- & $1.00\times$ & 4.24 & --- \\
\hspace{0.8em}Hard-Last-3 (default) & 60.5 $\pm$ 0.7 (12.1\%) & 60.9 & 24.8K & 1.51M & 92.5\% & 1.86M & $2.24\times$ & 2.10 & 1.64 \\
\midrule
\multicolumn{10}{@{}l}{\textit{Reduction baselines (same window and threshold)}} \\
Masking ($K{=}3$)~\citep{lindenbauer2025complexity} & 61.0 $\pm$ 5.3 (12.2\%) & 60.6 & 20.1K & 1.22M & 91.5\% & --- & $2.39\times$ & 1.69 & 0 \\
SWE-Pruner~\citep{wang2026swepruner} & 67 (13.4\%) & 48.5 & 29.1K & 1.41M & 94.0\% & 11.8K & $1.47\times$ & 1.83 & 0.01 \\
Self-Prune & 67 (13.4\%) & 50.4 & 38.8K & 1.96M & 94.3\% & 17.6K & $1.34\times$ & 2.72 & 0.13 \\
LongCodeZip~\citep{shi2025longcodezip} & 71 (14.2\%) & 43.5 & 38.3K & 1.67M & 96.1\% & 13.6K & $1.01\times$ & 1.62 & 0.04 \\
\bottomrule
\end{tabular}\end{adjustbox}
\end{table*}

\begin{table*}[t]
\centering
\caption{SWE-Master-4B-RL (131K context, 100 iterations). Reporting follows Table~\ref{tab:rq1}; each family uses its own uncompressed reference.}
\label{tab:transfer}
\scriptsize
\setlength{\tabcolsep}{2.5pt}
\begin{adjustbox}{max width=\textwidth}\begin{tabular}{@{}l r r r r r r r r r@{}}
\toprule
\textbf{Condition} & \textbf{resolved} & \textbf{calls} & \textbf{in/call} & \textbf{dec in} & \textbf{cached} & \textbf{enc/aux tok} & \textbf{ratio} & \textbf{dec PF} & \textbf{enc/aux PF} \\
\midrule
Uncompressed teacher & 137.0 $\pm$ 5.7 (27.5\%) & 85.0 & 51.8K & 4.40M & 97.2\% & --- & $1.00\times$ & 3.89 & --- \\
\midrule
\multicolumn{10}{@{}l}{\textit{LOHA + ACD (ours)}} \\
\hspace{0.8em}plain text ($K{=}\infty$) & 121.5 $\pm$ 6.4 (24.3\%) & 78.9 & 50.1K & 3.95M & 92.3\% & --- & $1.00\times$ & 6.77 & --- \\
\hspace{0.8em}Hard-Last-3 (default) & 109.0 $\pm$ 7.1 (21.8\%) & 81.5 & 22.2K & 1.81M & 85.0\% & 2.45M & $2.35\times$ & 4.34 & 2.16 \\
\midrule
Masking ($K{=}3$)~\citep{lindenbauer2025complexity} & 88.0 $\pm$ 1.0 (17.6\%) & 86.6 & 18.9K & 1.64M & 82.3\% & --- & $2.62\times$ & 4.22 & 0 \\
SWE-Pruner~\citep{wang2026swepruner} & 120 (24.0\%) & 82.8 & 36.3K & 3.00M & 86.9\% & 57.9K & $1.36\times$ & 7.75 & 0.07 \\
Self-Prune & 109 (21.8\%) & 76.4 & 33.8K & 2.58M & 92.0\% & 78.8K & $1.47\times$ & 4.13 & 0.57 \\
LongCodeZip~\citep{shi2025longcodezip} & 120 (24.0\%) & 79.3 & 48.6K & 3.85M & 91.5\% & 66.6K & $1.00\times$ & 7.60 & 0.17 \\
\bottomrule
\end{tabular}\end{adjustbox}
\end{table*}

\textbf{Qwen3-4B-Instruct.}
LOHA + ACD resolves 12.1\% of instances against 14.5\% for the uncompressed base and 13.2\% for the adapted full-text agent. Mean context per call falls from the base agent's 43.6K to 24.8K tokens. Masking reaches a similar resolve rate (12.2\%) with 20.1K tokens per call. The pruning baselines resolve 13.4--14.2\% while retaining 29--39K tokens per call.

\phantomsection\textbf{SWE-Master-4B-RL.}
\label{sec:eval:transfer}
LOHA + ACD resolves 21.8\% against 27.5\% for the uncompressed base and 24.3\% for the adapted full-text agent, while reducing mean context per call from 51.8K to 22.2K tokens. Masking drops to 17.6\%; retaining a latent history therefore preserves more task performance on this base. SWE-Pruner and LongCodeZip resolve 24.0\%, and Self-Prune matches LOHA's 21.8\%, with all three retaining larger contexts (34--49K tokens per call).

\textbf{Trajectory cost.}
Per-call savings translate differently across the two agents. Qwen3 with LOHA makes more calls than its base (60.9 versus 37.9), so decoder input falls only from 1.65M to 1.51M tokens per trajectory. SWE-Master makes a similar number of calls (81.5 versus 85.0), and decoder input falls from 4.40M to 1.81M. The representation ratios, measured against each agent's own contexts rendered in full text, are $2.24\times$ and $2.35\times$, respectively.

\subsection{Performance under Context Limits}
\label{sec:eval:window}

\begin{figure}[t]
\centering
\includegraphics[width=\textwidth]{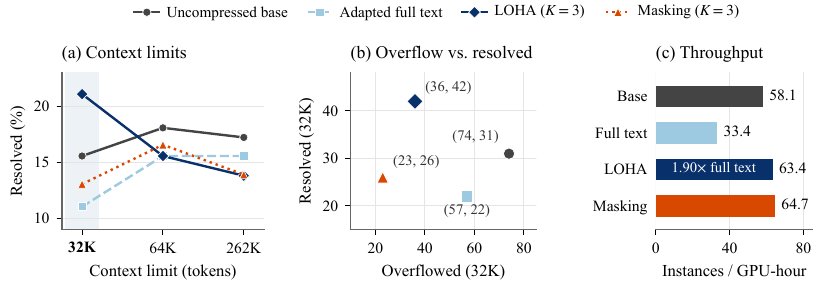}
\caption{Qwen3: (a) resolve rates on 199 instances: single runs at 32K/64K, seed means at 262K. (b) Overflow and resolved counts at 32K. (c) Single-GPU throughput at concurrency 16. Protocols and paired tests: Appendix~\ref{app:results}.}
\label{fig:window}
\end{figure}

For Qwen3, we repeat the comparison with 64K- and 32K-token context limits on a 199-instance subset. A trajectory that exceeds the limit is stopped and its patch is graded. At 64K, the four conditions resolve 31--36 instances. At 32K, Hard-Last-3 resolves 42, compared with 22 for the adapted full-text agent, 26 for masking, and 31 for the base agent (Figure~\ref{fig:window}a). The gains over the adapted full-text agent and masking are supported by paired McNemar tests ($p{=}0.0002$ and $p{=}0.0025$, respectively).

Masking exceeds the 32K limit less often than LOHA (23 versus 36 trajectories) but resolves fewer tasks (Figure~\ref{fig:window}b). Fitting within the context window is therefore only part of the benefit: the results also favor retaining a compressed history over deleting older observations. Full counts and paired comparisons are in Appendix~\ref{app:results:window}.

\subsection{Ablation Studies}
\label{sec:eval:ablations}

We vary the recency window for both model families and the distillation teacher for Qwen3. The teacher ablation uses rank-64 students trained on 3{,}514 trajectories.

\begin{table}[t]
\centering
\caption{Recency and teacher ablations. (a) 499 instances per model; dec PF: estimated decoder PFLOPs per trajectory. (b) 199 random instances with the teacher supplying both targets, followed by a separate 20-task diagnostic listing teacher/anchor pairs. Res.: resolved; patch: non-empty patches; stuck: loop terminations. Single runs except the 199-task base mean. Details: Appendices~\ref{app:seeds} and~\ref{app:results:rq2}.}
\label{tab:sweeps-main}
\label{tab:rq2}
\footnotesize
\begin{minipage}[t]{0.35\linewidth}
\centering
\textbf{(a) Recency-window sweep}\par\smallskip
\setlength{\tabcolsep}{3pt}
\begin{tabular*}{\linewidth}{@{\extracolsep{\fill}}c rr rr@{}}
\toprule
& \multicolumn{2}{c}{Qwen3} & \multicolumn{2}{c}{SWE-Master} \\
\cmidrule(lr){2-3}\cmidrule(l){4-5}
$K$ & res. & dec PF & res. & dec PF \\
\midrule
$\infty$ & 80 & 1.47 & 128 & 2.99 \\
8 & 72 & 2.10 & 115 & 8.37 \\
4 & 76 & 2.08 & 108 & 4.49 \\
\textbf{3} & 68 & 1.67 & 111 & 3.74 \\
2 & 65 & 1.40 & 103 & 2.93 \\
1 & 47 & 1.21 & 88 & 1.89 \\
0 & 37 & 1.18 & 51 & 0.93 \\
\bottomrule
\end{tabular*}
\end{minipage}\hfill
\begin{minipage}[t]{0.62\linewidth}
\centering
\textbf{(b) Distillation teacher}\par\smallskip
\setlength{\tabcolsep}{3pt}
\begin{tabular*}{\linewidth}{@{\extracolsep{\fill}}l rrr rrr@{}}
\toprule
& \multicolumn{3}{c}{Full text} & \multicolumn{3}{c}{Hard-Last-3} \\
\cmidrule(lr){2-4}\cmidrule(l){5-7}
Teacher & res. & patch & stuck & res. & patch & stuck \\
\midrule
Uncompressed base & 34.3 & --- & --- & --- & --- & --- \\
\textbf{Self (Qwen3-4B)} & \textbf{28} & 140 & 43 & \textbf{29} & 127 & 50 \\
Qwen3-30B-A3B & 15 & 68 & 136 & 14 & 59 & 141 \\
SWE-Master-4B-RL & 0 & 22 & 59 & 0 & 22 & 68 \\
\midrule
\multicolumn{7}{@{}l}{\textit{20 tasks: distillation teacher / anchor}} \\
Uncompressed base & 10/20 & 14 & 2 & --- & --- & --- \\
Self / self & 11/20 & 14 & 3 & 11/20 & 16 & 5 \\
30B / 30B & 7/20 & 8 & 10 & 6/20 & 8 & 9 \\
30B / self & 12/20 & 15 & 5 & 9/20 & 13 & 6 \\
\bottomrule
\end{tabular*}
\end{minipage}
\end{table}

\begin{samepage}
\phantomsection
\label{sec:eval:rq3}
\textbf{Recency-Window Sweep.}
We sweep $K\in\{0,1,2,3,4,8\}$ and full text while holding each trained student fixed (Table~\ref{tab:sweeps-main}a). Larger windows generally improve resolve rate while retaining more observations in text. Both models lose performance at $K{=}1$ and again at $K{=}0$; a separate 20-task diagnostic finds that increasing patch rank does not repair the fully latent policy (Appendix~\ref{app:results:rq3}).
\par
\end{samepage}

\begin{samepage}
In this single-run sweep, increasing $K$ from 3 to 8 raises resolved counts from 68 to 72 on Qwen3 (13.6\% to 14.4\%) and from 111 to 115 on SWE-Master (22.2\% to 23.0\%). Estimated decoder compute rises from 1.67 to 2.10 PF and from 3.74 to 8.37 PF per trajectory, respectively. The trend is not strictly monotonic: Qwen3 resolves 76 tasks at $K{=}4$. Thus, $K{=}3$ is a compact default, while $K{=}8$ offers higher resolve rates at greater text retention and compute cost.
\par
\end{samepage}

On 358 literals quoted by later actions, the released LCLM decoder achieves exact-recitation rates of 20.4\%, 15.6\%, and 15.1\% at $4\times$, $8\times$, and $16\times$ compression, respectively (Appendix~\ref{app:recitation}). The low fidelity even at $4\times$ supports retaining recent source text for exact tool arguments.

\phantomsection
\label{sec:eval:rq2}
\textbf{Distillation Teacher.}
The 199-task block of Table~\ref{tab:rq2}b uses the distillation teacher for both targets. Under full text and Hard-Last-3, the self-teacher resolves 28 and 29 tasks; the 30B teacher resolves 15 and 14 with more loop terminations, and SWE-Master resolves none. On the same 20 diagnostic tasks, the rank-64 self-teacher resolves 11 tasks under either policy; using the 30B model for both targets resolves seven and six. Keeping the base anchor with the 30B distillation teacher gives 12 and nine, with forward KL for both terms. The latter uses top-64 rather than top-16 cached targets. These diagnostics support retaining the base anchor while allowing a different distillation teacher.

Better literal recall need not improve task performance. At rank 256, the student using the 30B model for both targets exactly recites 35/96 literals, versus 26/96 for the self-anchored student, but resolves only 4/20 tasks versus 10/20 under either context policy. Full controls and KL-direction ablations are in Appendix~\ref{app:results:rq2}.

\textbf{Reading Initialization.}
On Qwen3, the rank-64 initialization retains 17\% of the LCLM--base weight-difference energy yet recovers 79\% of the LCLM decoder's structured read-back score. Before ACD training, this patch still degrades agent behavior (Appendix~\ref{app:mechanism}).

\subsection{Serving Efficiency}
\label{sec:eval:rq4}

For Qwen3, on one GPU with 16 concurrent trajectories, LOHA completes 63.4 instances per GPU-hour, compared with 33.4 for the adapted full-text agent and 58.1 for the base agent (Figure~\ref{fig:window}c). This is a $1.9\times$ throughput increase over the adapted full-text condition, with median call latency falling from 5.07 to 1.71 seconds and 90th-percentile latency from 45.6 to 7.17 seconds. Masking reaches 64.7 instances per GPU-hour. At the measured mean context lengths, the shared 416K-token KV cache accommodates about 11 adapted full-text trajectories versus 20 with Hard-Last-3, below and above the concurrency of 16, respectively. Full serving measurements are in Appendix~\ref{app:results:rq4}.

In the current server, Qwen3 with LOHA uses an estimated 2.10 PF of decoder compute and 1.64 PF of encoder compute per trajectory, against 1.82 PF for the base agent. The server re-encodes old observations at every call. A replay with per-observation caching reduces the encoder contribution to 0.02 PF and estimates total compute at 2.12 PF per trajectory. For SWE-Master, estimated compute is 4.34 PF in the decoder and 2.16 PF in the encoder, against 3.89 PF for its uncompressed base.

\section{Related Work}
\label{sec:related-work}

\textbf{Context and code compression.}
Agent histories can be shortened by masking~\citep{lindenbauer2025complexity}, summarization~\citep{packer2023memgpt,kang2025acon}, or folding~\citep{sun2025contextfolding,ye2025agentfold}. SWE-Pruner and SWE-Pruner Pro select observation lines~\citep{wang2026swepruner,wang2026sweprunerpro}; CoACT uses action-preserving targets~\citep{chen2026coact}, while TACO evolves output-filtering rules~\citep{ren2026taco}. Prompt compressors select tokens or spans~\citep{li2023selectivecontext,jiang2023llmlingua,pan2024llmlingua2}, and LongCodeZip and SlimCode exploit code structure~\citep{shi2025longcodezip,wang2024slimcode}. LOHA instead keeps older observations in latent form and recent content verbatim. Masking provides a direct comparison by using the same recency window without retaining a latent history.

\textbf{Soft-token compression and memory.}
Gist tokens, in-context autoencoders, and extreme compressors establish the soft-token interface~\citep{mu2023gist,ge2024icae,li2024500xcompressor}; LCLM scales it to long documents~\citep{li2026lclm}. Latent memory systems target experience reuse and agent coordination~\citep{zhang2026memw,zhang2025memgen,fu2026latentmem}. Our setting instead requires compressed observations to support repeated tool interactions within a trajectory, including actions that need exact text.

\textbf{Distillation and anchoring.}
Context distillation transfers predictions from richer to reduced inputs~\citep{askell2021general,snell2022context,cheng2024xrag}. Behavioral anchoring relates to Learning without Forgetting~\citep{li2018lwf}, self-distillation~\citep{yang2024sdft}, and reference-policy KL regularization~\citep{ziegler2019finetuning,ouyang2022training}. ACD combines these roles: the same base model supplies the latent-view target and full-text reference, while a low-rank patch~\citep{hu2022lora} enables adaptation. Our ablations distinguish the effects of the distillation teacher and behavioral anchor. Extended comparisons are in Appendix~\ref{app:related}.

\section{Conclusion and Future Works}
\label{sec:conclusion}

LOHA and ACD combine observation compression with behavior-preserving adaptation for software-engineering agents. LOHA keeps recent observations and agent turns in text while compressing older observations; ACD teaches latent reading while anchoring full-text behavior to the base policy. At $K{=}3$, context per call falls by 43\% on Qwen3 and 57\% on SWE-Master. Larger recency windows trade compression for task performance, while the 32K experiments demonstrate the value of retaining compressed history. The teacher ablations further motivate treating context representation and policy adaptation as a joint design problem.

Future work can pursue three directions. First, learning to allocate observation fidelity by anticipated tool use could jointly determine what to compress and when to retrieve exact source text. Second, objectives that constrain tool-use decisions or trajectory-level outcomes, rather than only next-token distributions, could offer new ways to balance latent-reading gains against policy drift. Third, training on the adapted agent's own trajectories could couple representation learning to the actions and observations it induces, with task success guiding what information the compressed history preserves. Together, these directions would connect compression more closely to how an agent uses its context.

\clearpage
\bibliography{refs}
\bibliographystyle{iclr2027_conference}

\clearpage
\appendix

\section{Experimental Protocol}
\label{app:protocol}

The main comparison evaluates LOHA + ACD and the same compression baselines on Qwen3-4B-Instruct-2507 and SWE-Master-4B-RL. This section specifies the common scaffold, each model family's protocol, and the reported metrics. Additional results are in Appendix~\ref{app:results}; implementation and training details are in Appendix~\ref{app:method}.

\subsection{Benchmark, Agents, and Decoding}

We evaluate on 499 of the 500 SWE-bench Verified instances~\citep{jimenez2024swebench}. The instance \texttt{django\_\_django-12273} is excluded from every condition because its official evaluation image fails during container initialization (\texttt{/usr/bin/cat} is missing, exit code 127).

Both model families use OpenHands~\citep{wang2024openhands} 0.62.0 with CodeActAgent and native function calling. The tools are \texttt{execute\_bash}, \texttt{str\_replace\_editor}, \texttt{think}, \texttt{finish}, and \texttt{task\_tracker}; browsing, Jupyter, and the history condenser are disabled. Qwen3 uses a 262{,}144-token context window and a 200-iteration cap. SWE-Master uses its native tool rendering, a 131{,}072-token window, and a 100-iteration cap. Conditions within each family share decoding settings; for Qwen3 these are temperature 0.7, top-$p$ 0.8, and top-$k$ 20, enforced in both the server and harness.

Each instance contributes one final result per seed. The main evaluation uses a 7{,}200-second wall-clock limit; an in-process alarm at 6{,}600 seconds ends the trajectory and extracts its patch. We run three seeds for the Qwen3 base and masking, and two for its ACD conditions. For SWE-Master, we run two seeds for the base and ACD conditions and three for masking. Pruning baselines use one run per family. Per-seed counts and recency-window sweeps are listed in Appendix~\ref{app:seeds}.

\subsection{Conditions and Baselines}

Each model family has its own uncompressed reference. The ACD student is initialized from that model and uses it as the teacher for both objectives. Comparing the student in full text with the base measures adaptation; comparing the same student in full text and under LOHA measures the change in context representation.

\begin{itemize}[leftmargin=1.2em,itemsep=3pt,topsep=3pt]
  \item \textbf{Uncompressed base.} The original Qwen3 or SWE-Master agent with full-text observations.
  \item \textbf{LOHA + ACD.} The rank-64 self-anchored student, evaluated with Hard-Last-3 and full text ($K{=}\infty$) in the main comparisons. Recency sweeps test $K\in\{0,1,2,3,4,8\}$ for both model families; the 20-instance diagnostic also tests rank 256.
  \item \textbf{Observation masking}~\citep{lindenbauer2025complexity}. The last three observations remain in text, and older observations are removed. This shares LOHA's recent text window and tests the value of retaining a latent history.
  \item \textbf{SWE-Pruner}~\citep{wang2026swepruner}. A 0.6B reranker selects observation lines using a focus question derived from the agent's last action.
  \item \textbf{Self-Prune.} The 4B decoder selects numbered lines through guided JSON, with at most 400 lines per pruning call. It uses the same interface as SWE-Pruner without a separate model.
  \item \textbf{LongCodeZip}~\citep{shi2025longcodezip}. A 1.5B ranker selects code chunks and compresses observations in token space.
\end{itemize}

All compression baselines use the same $K{=}3$ recent-observation window and 128-character threshold. The pruning servers cache the compressed text and process each observation once. The current LOHA server re-encodes compressed observations at each model call; the measured cost and a replay with encoder caching are reported separately in Appendix~\ref{app:results:rq4}.

\subsection{Scaffold Changes and Grading}

The shared scaffold fork renders observations according to the selected context policy. Observations shorter than 128 characters remain in text under every inference policy. The editor also supports a bounded fallback when a \texttt{str\_replace} argument does not match exactly: it first normalizes whitespace line by line, then tries a semi-global edit-distance match accepted only when unique by a margin. A syntax check precedes writing. Fallback events are logged. The Qwen3 comparisons also share three editor hints for unchanged files, missing parameters, and unresolved paths.

Containers run rootless on the official SWE-bench images through \texttt{udocker}'s \texttt{fakechroot} engine. Host paths can appear in observation output because the engine does not rewrite those strings; the inspected tool arguments address \texttt{/workspace/\ldots}. This behavior is shared across conditions.

Grading uses the official SWE-bench harness, version 4.1.0. We read its complete log from the container file system because truncation in the scaffold's observation layer can remove the start-of-test marker and misclassify a passing patch as unapplied. Gold patches were used to validate this grading path. Empty predictions count as unresolved.

\subsection{Metrics}

\textbf{Task performance and behavior.}
Resolve rate is the fraction of the 499 instances that pass the benchmark grader. In the ablation tables, \emph{res.} is the resolved count, \emph{patch} counts non-empty patches touching a file other than the build manifest, \emph{stuck} counts loop terminations, and \emph{cap} counts iteration-limit terminations. The \emph{edits} and \emph{calls} columns are mean \texttt{str\_replace} edits and model calls per trajectory. We also log tool-call validity and editor-fallback use.

\textbf{Token cost.}
The \emph{in/call} column is mean decoder prompt length; \emph{dec in} sums prompt tokens over a trajectory; \emph{cached} is the share served from the prefix cache. The \emph{enc/aux tok} column separately counts tokens processed by the observation encoder or baseline compressor. Self-Prune's auxiliary work uses the decoder itself.

The \emph{ratio} column divides the plaintext equivalent of an agent's own contexts by the decoder tokens actually read. It measures the representation reduction for those contexts. Actual token savings between agents also depend on the trajectories they generate. Encoded observation bodies have a $16\times$ ratio, while prompts, assistant turns, tool envelopes, and recent observations remain in text.

\textbf{Compute and serving.}
We estimate compute per trajectory in petaflops (\emph{PF}), using $2P$ FLOPs per computed token plus the quadratic attention term, with cached positions excluded from prefill. The decoder has $P{=}3.63$B non-embedding parameters; encoder and auxiliary models use dense-only estimates at their respective sizes. Serving measurements report completed instances per GPU-hour, trajectory wall time, and per-call latency. Their workload and completion cutoff are given in Appendix~\ref{app:results:rq4}.

\textbf{Reading probes.}
We measure exact recitation of 96 literal fields and structured extraction of 48 fields from compressed tool outputs. The structured probe uses exact-JSON and substring scoring to distinguish formatting errors from content recovery. These adaptation probes are separate from the 358-literal compression study in Appendix~\ref{app:recitation}.

\section{Additional Results}
\label{app:results}

This section supplements the two main result tables with paired comparisons, context-limit experiments, ablations, and serving measurements. Unless stated otherwise, the ablations and serving experiments use the Qwen3 family.

\subsection{Full-Window Comparisons}
\label{app:results:rq1}

\textbf{Qwen3-4B-Instruct.}
For Table~\ref{tab:rq1}, paired McNemar tests against the base agent give $p{=}0.19$ and $1.00$ for the adapted full-text condition, and $p{=}0.049$ and $0.32$ for Hard-Last-3. In the first Hard-Last-3 comparison, 18 instances are resolved only by the adapted agent and 33 only by the base. The two Hard-Last-3 seeds have representation ratios of $2.20\times$ and $2.27\times$.

Masking has the largest variation across seeds and averages 99 iteration-cap terminations against 43 for the base. The observation compression ratios are $1.02\times$ for LongCodeZip, $2.9\times$ for SWE-Pruner, and $1.8\times$ for Self-Prune, compared with LOHA's $16\times$. LongCodeZip retains nearly every chunk in its coarse ranking stage on these observations.

\phantomsection\textbf{SWE-Master-4B-RL.}
\label{app:results:transfer}
For Table~\ref{tab:transfer}, paired tests against the SWE-Master base give $p{=}0.13$ and $0.12$ for the adapted full-text condition, $p{=}0.07$ and $p{<}0.001$ for Hard-Last-3, and $p{<}0.001$ in both comparisons for which masking and base runs are paired. The mean resolve-rate gaps are 3.2 points for adapted full text, 5.7 for Hard-Last-3, and 9.9 for masking.

SWE-Pruner and LongCodeZip each resolve 120 instances, and Self-Prune resolves 109, compared with a mean of 109 for LOHA. Their contexts remain larger: 34--49K tokens per call against LOHA's 22.2K. Measured observation ratios are $1.7\times$, $1.9\times$, and $1.01\times$ for SWE-Pruner, Self-Prune, and LongCodeZip, respectively. All comparisons in this family use SWE-Master as the starting agent and, for ACD, as its own anchor.

\subsection{Context-Limit Experiments}
\label{app:results:window}

Table~\ref{tab:window} evaluates the four Qwen3 conditions on the same 199-instance subset with context limits of 65{,}536 and 32{,}768 tokens, reserving 4{,}096 tokens for generation. Each capped condition is run once. The 262K column is the mean from the full-window seeds on this subset. When a trajectory exceeds the limit, generation stops and the current patch is graded.

\begin{table}[t]
\centering
\caption{Resolve rate under a capped context window on a 199-instance subset, one run per cell; the 262K column is the mean over the seeds of Table~\ref{tab:rq1} on the same subset. \emph{ovf}: trajectories stopped at the cap. Paired tests in the text.}
\label{tab:window}
\footnotesize
\setlength{\tabcolsep}{4pt}
\begin{adjustbox}{max width=\linewidth}\begin{tabular}{@{}l r rr rr@{}}
\toprule
& \textbf{262K} & \multicolumn{2}{c}{\textbf{64K}} & \multicolumn{2}{c}{\textbf{32K}} \\
\cmidrule(lr){3-4}\cmidrule(l){5-6}
\textbf{Condition} & res. & res. & ovf & res. & ovf \\
\midrule
Uncompressed anchor & 34.3/199 & 36/199 & 24 & 31/199 & 74 \\
LOHA plain text ($K{=}\infty$) & 31.0/199 & 31/199 & 25 & 22/199 & 57 \\
LOHA Hard-Last-3 & 27.5/199 & 31/199 & 13 & 42/199 & 36 \\
Masking ($K{=}3$) & 27.7/199 & 33/199 & 7 & 26/199 & 23 \\
\bottomrule
\end{tabular}\end{adjustbox}
\end{table}

At 32K, the paired comparison between Hard-Last-3 and adapted full text has 24 instances resolved only by Hard-Last-3 and four only by full text ($p{=}0.0002$). Against masking the counts are 21 and five ($p{=}0.0025$); against the base they are 19 and eight ($p{=}0.052$). Of the overflowed trajectories, four base runs, zero adapted full-text runs, one Hard-Last-3 run, and two masking runs still yield passing patches. Masking overflows less often than Hard-Last-3 but resolves fewer tasks, supporting the value of retaining compressed history.

\subsection{Recency Ablation}
\label{app:results:rq3}

The 20-instance recency diagnostic and the teacher and objective ablations use Qwen3 students trained on the first 3{,}514 trajectories in a fixed corpus ordering. The diagnostic subset contains 20 instances drawn from the 69 solved by one full-benchmark Qwen3 base run. The base solves 10 of these on re-evaluation. This subset is used to examine behavioral failures, and its resolve rates are reported separately from the main benchmark results.

\begin{table}[t]
\centering
\caption{Behavior and cost of the self-anchored model under the three settings of the recency policy, on the 20-instance subset of Table~\ref{tab:rq2-full}. \emph{dec tok}, \emph{enc tok}: decoder prompt and encoder input tokens per trajectory; other columns as in Appendix~\ref{app:protocol}.}
\label{tab:rq3}
\small
\setlength{\tabcolsep}{4pt}
\begin{adjustbox}{max width=\linewidth}\begin{tabular}{@{}l c rrrr rrrr@{}}
\toprule
\textbf{model} & \textbf{policy} & res. & patch & stuck & cap & dec tok & cached & enc tok & ratio \\
\midrule
Uncompressed agent  & plain            & 10/20 & 14 & 2 & 1 & 846K  & 98.0\% & ---   & $1.00\times$ \\
\midrule
ACD, $\rho{=}64$    & $K{=}\infty$     & 11/20 & 14 & 3 & 4 & 2.39M & 99.2\% & ---   & $1.00\times$ \\
ACD, $\rho{=}64$    & \textbf{$K{=}3$} & \textbf{11/20} & 16 & 5 & 1 & 632K  & 92.1\% & 763K  & $2.21\times$ \\
ACD, $\rho{=}64$    & $K{=}0$          & 1/20  & 4  & 5 & 9 & 2.30M & 99.6\% & 4.10M & $2.78\times$ \\
\midrule
ACD, $\rho{=}256$   & $K{=}\infty$     & 10/20 & 15 & 3 & 1 & 1.74M & 98.4\% & ---   & $1.00\times$ \\
ACD, $\rho{=}256$   & $K{=}3$          & 10/20 & 16 & 2 & 2 & 505K  & 95.4\% & 188K  & $1.37\times$ \\
ACD, $\rho{=}256$   & $K{=}0$          & 2/20  & 3  & 5 & 8 & 3.17M & 99.7\% & 6.98M & $3.20\times$ \\
\bottomrule
\end{tabular}\end{adjustbox}
\end{table}

At rank 64, Hard-Last-3 and full text both resolve 11 tasks; at rank 256, both resolve 10. Hard-Last-3 produces 16 non-empty patches at either rank, compared with 14 and 15 for full text. Fully latent observations reduce these counts to one or two resolved tasks and three or four patches.

The fully latent agents remain active, averaging 18--26 edits and 94--103 model calls per trajectory, with eight or nine runs reaching the iteration cap. Their trajectories include reconstructed edit strings, inaccurate test commands, and repeated views that return another compressed observation. Keeping recent observations in text addresses this failure mode on the diagnostic subset.

\subsection{Anchor and Objective Ablations}
\label{app:results:rq2}

\textbf{Reference model and patch rank.}
Table~\ref{tab:rq2-full} varies the reference for both ACD objectives and the patch rank. The upper panel uses the 20-instance subset above. The external SWE-Master-anchor runs were stopped early after producing no patches; their completed sample sizes are shown. The lower panel evaluates the three rank-64 students on a 199-instance random subset, selected by a seed-0 shuffle, with one run per condition.

\begin{table}[t]
\centering
\caption{Anchor ablation. Each row is one ACD run, varying the model from which both terms distill and the patch rank $\rho$, evaluated in plain text ($K{=}\infty$) and under Hard-Last-3. Upper panel: the 20-instance subset described in the text; SWE-Master rows were stopped early at zero patches and show their $n$. Lower panel: the three $\rho{=}64$ students on a 199-instance subset, one run each, with the untrained agent's seed mean as reference. Column definitions in Appendix~\ref{app:protocol}.}
\label{tab:rq2-full}
\footnotesize
\setlength{\tabcolsep}{3.5pt}
\begin{adjustbox}{max width=\linewidth}\begin{tabular}{@{}l c rrrrr rrrrr@{}}
\toprule
& & \multicolumn{5}{c}{\textbf{plain text ($K{=}\infty$)}} & \multicolumn{5}{c}{\textbf{Hard-Last-3}} \\
\cmidrule(lr){3-7}\cmidrule(l){8-12}
\textbf{anchor} & $\rho$ & res. & patch & stuck & edits & calls & res. & patch & stuck & edits & calls \\
\midrule
Uncompressed agent (no patch)      & --- & 10/20 & 14 & 2  & 4.8  & 23.7 & \multicolumn{5}{c}{---} \\
\midrule
\textbf{itself} (Qwen3-4B-Instruct) & 64  & \textbf{11/20} & 14 & 3  & 8.5  & 51.1 & \textbf{11/20} & 16 & 5 & 6.8 & 30.5 \\
itself                              & 256 & 10/20 & 15 & 3  & 4.1  & 25.4 & 10/20 & 16 & 2 & 1.5 & 30.8 \\
\midrule
Qwen3-30B-A3B (same family, larger) & 64  & 7/20  & 8  & 10 & 0.4  & 65.0 & 6/20  & 8  & 9 & 6.3 & 67.9 \\
Qwen3-30B-A3B                       & 256 & 4/20  & 5  & 15 & 1.0  & 41.4 & 4/20  & 6  & 12 & 0.8 & 50.8 \\
\midrule
SWE-Master-4B-RL (fine-tuned agent) & 64  & 0/8   & 0  & 7  & 0.0  & 49.0 & 0/5   & 0  & 5 & 0.0 & 30.6 \\
SWE-Master-4B-RL                    & 256 & 0/5   & 0  & 4  & 0.0  & 116.2 & 0/2  & 0  & 2 & 0.0 & 6.5 \\
\midrule
\multicolumn{12}{@{}l}{\textit{Same three $\rho{=}64$ students on a 199-instance subset (seed-0 shuffle of Verified; cap = 200-iteration cap)}} \\
\textbf{anchor} & $\rho$ & res. & patch & stuck & cap & calls & res. & patch & stuck & cap & calls \\
\midrule
Uncompressed agent (no patch) & --- & 34.3/199 & --- & --- & --- & --- & \multicolumn{5}{c}{---} \\
\textbf{itself} (Qwen3-4B-Instruct) & 64 & 28/199 & 140 & 43 & 24 & 43.2 & 29/199 & 127 & 50 & 34 & 54.3 \\
Qwen3-30B-A3B & 64 & 15/199 & 68 & 136 & 17 & 36.5 & 14/199 & 59 & 141 & 22 & 44.0 \\
SWE-Master-4B-RL & 64 & 0/199 & 22 & 59 & 15 & 24.2 & 0/199 & 22 & 68 & 7 & 20.6 \\
\bottomrule
\end{tabular}\end{adjustbox}
\end{table}

Self-anchored students retain task performance at both ranks on the small subset. Using the 30B reference increases loop terminations and reduces patch production; increasing rank from 64 to 256 does not repair this behavior. The larger subset confirms the ordering: self-anchoring resolves 28 tasks in full text and 29 under Hard-Last-3, the 30B reference resolves 15 and 14, and the external SWE-Master reference resolves none. The SWE-Master-reference runs produce 22 patches under each policy and reach the two-hour limit in 46 of 398 trajectories.

Reading scores do not follow agent performance. A rank-256 student using the 30B reference recites 35 of 96 literals exactly, compared with 26 for a self-anchored student, despite its worse benchmark behavior. The self-anchored subset-trained model also has resolved counts close to the full-data model on the 199-instance subset: 28 versus 31.0 in full text, and 29 versus 27.5 under Hard-Last-3.

\textbf{Distillation teacher and KL direction.}
Table~\ref{tab:rq2b} keeps the behavioral anchor at the base model while varying the distillation teacher and each KL direction. The seven variants use top-64 cached teacher distributions; the reference row uses top-16 distributions. All use rank 64 and the same 20-instance subset.

\begin{table}[t]
\centering
\caption{Direction of each divergence and teacher of the distillation term, with the anchor fixed to the base model ($\rho{=}64$; subset and columns as in Table~\ref{tab:rq2-full}). F: forward KL; R: reverse KL. The reference row is the self-anchored run of Table~\ref{tab:rq2-full}.}
\label{tab:rq2b}
\footnotesize
\setlength{\tabcolsep}{2.8pt}
\begin{adjustbox}{max width=\linewidth}\begin{tabular}{@{}ccc rrrrr rrrrr@{}}
\toprule
& & & \multicolumn{5}{c}{\textbf{plain text ($K{=}\infty$)}} & \multicolumn{5}{c}{\textbf{Hard-Last-3}} \\
\cmidrule(lr){4-8}\cmidrule(l){9-13}
\textbf{anchor KL} & \textbf{distill.\ teacher} & \textbf{distill.\ KL} & res. & patch & stuck & edits & calls & res. & patch & stuck & edits & calls \\
\midrule
F & itself (4B) & F (ref.; top-16) & 11/20 & 14 & 3 & 8.5 & 51.1 & 11/20 & 16 & 5 & 6.8 & 30.5 \\
\midrule
F & itself (4B) & R & 10/20 & 13 & 4  & 4.4 & 40.8 & 9/20  & 12 & 4 & 17.0 & 68.8 \\
R & itself (4B) & F & 11/20 & 13 & 4  & 9.1 & 34.3 & 6/20  & 10 & 9 & 9.2  & 44.5 \\
R & itself (4B) & R & 9/20  & 14 & 3  & 8.6 & 42.6 & 8/20  & 13 & 8 & 1.3  & 13.9 \\
\midrule
F & Qwen3-30B-A3B & F & 12/20 & 15 & 5  & 7.1 & 37.7 & 9/20  & 13 & 6 & 1.9  & 37.6 \\
F & Qwen3-30B-A3B & R & 5/20  & 8  & 10 & 3.9 & 23.2 & 7/20  & 10 & 9 & 4.5  & 48.4 \\
R & Qwen3-30B-A3B & F & 8/20  & 13 & 7  & 5.4 & 27.2 & 8/20  & 14 & 5 & 4.0  & 50.2 \\
R & Qwen3-30B-A3B & R & 10/20 & 14 & 6  & 4.4 & 23.6 & 10/20 & 12 & 7 & 8.9  & 43.4 \\
\bottomrule
\end{tabular}\end{adjustbox}
\end{table}

Six variants resolve 8--12 tasks in full text and 6--10 under Hard-Last-3. The remaining variant, a forward-KL anchor with reverse-KL distillation from the 30B teacher, resolves five in full text and has 10 loop terminations. Fully latent evaluation resolves only 1--3 tasks in every variant. There is no consistent advantage from reversing either KL direction or using the larger distillation teacher. Compared with replacing both references, retaining the base-model anchor avoids the widespread collapse in most variants.

\textbf{Cross-entropy variant.}
An earlier experiment replaced latent-view distillation with cross-entropy on trajectory tokens while keeping the anchor loss. Under a preliminary harness, this variant improved structured extraction from 17 to 24 fields and exact recitation from 22 to 28, but edited no files across eight plain-text tasks, emitted 15 malformed tool calls, and never called \texttt{str\_replace}. The complete objective solved five of those eight tasks. This preliminary result illustrates why the behavioral probe accompanies the reading probes.

\subsection{Serving Measurements and Encoder Cost}
\label{app:results:rq4}

\textbf{Workload.}
Table~\ref{tab:throughput} uses one RTX PRO 6000 GPU with 96 GB of memory per condition, the same 100 Qwen3 evaluation instances, and 16 concurrent trajectories. Throughput is measured through the 84th completion to exclude the final tail of long trajectories. Wall time and latency columns report the median and 90th percentile.

\begin{table}[t]
\centering
\caption{Serving throughput on one GPU per condition (RTX PRO 6000, 96 GB), 100 instances at 16 concurrent trajectories. \emph{inst/GPU-h}: completions per GPU-hour up to the 84th, before the tail of long trajectories; \emph{wall}: trajectory wall-clock, median and 90th percentile (minutes); \emph{latency}: per-call decoder latency, median and 90th percentile (seconds); \emph{KV}: KV-cache capacity in tokens; \emph{adm.}: full-context trajectories it admits at in/call.}
\label{tab:throughput}
\footnotesize
\setlength{\tabcolsep}{3.5pt}
\begin{adjustbox}{max width=\linewidth}\begin{tabular}{@{}l r r r r r r@{}}
\toprule
\textbf{Condition} & \textbf{inst/GPU-h} & \textbf{wall (min)} & \textbf{latency (s)} & \textbf{in/call} & \textbf{KV} & \textbf{adm.} \\
\midrule
Uncompressed anchor & 58.1 & 9.1 / 21.0 & 2.04 / 9.99 & 35.6K & 547K & 15.4 \\
LOHA plain text ($K{=}\infty$) & 33.4 & 9.4 / 85.4 & 5.07 / 45.6 & 38.1K & 416K & 10.9 \\
LOHA Hard-Last-3 & 63.4 & 8.5 / 17.7 & 1.71 / 7.17 & 20.8K & 416K & 20.0 \\
Masking ($K{=}3$) & 64.7 & 8.3 / 20.4 & 1.44 / 7.13 & 19.9K & 416K & 20.9 \\
\bottomrule
\end{tabular}\end{adjustbox}
\end{table}

The serving shim keeps the encoder resident and provides 416K tokens of KV-cache capacity, compared with 547K for the base agent served by plain vLLM. At the measured mean context lengths, this accommodates about 20 Hard-Last-3 trajectories, 11 adapted full-text trajectories, and 15 base-agent trajectories. The adapted full-text condition exceeds that capacity at concurrency 16 and has a long latency tail. Masking and LOHA have similar context sizes and throughput.

\textbf{Prefix caching.}
In the 20-instance recency ablation, the fully latent layout reuses 99.6--99.7\% of decoder prompt tokens from the prefix cache, full text reuses 98.4--99.2\%, and Hard-Last-3 reuses 92.1--95.4\%. The main Qwen3 and SWE-Master tables report Hard-Last-3 cache shares of 92.5\% and 85.0\%, respectively. These are measurements of their respective workloads; the cost of advancing the recent window depends on the trajectories.

\textbf{Encoder replay.}
The current server encodes every compressed observation again at each call. We replay Qwen3 Hard-Last-3 trajectories to compare this implementation with a cache keyed by observation text. The replay reproduces processed encoder tokens with a median ratio of 0.985 over 490 trajectories. Caching reduces encoder input from 1.87M to 25.7K tokens per trajectory, or from 1.65 PF to 0.02 PF. Deduplicating content-identical observations reduces it further to 11.0K tokens. A second seed gives 1.95M to 26.7K; an additional run gives 1.31M to 21.7K.

With per-observation caching, the replay estimates total LOHA compute at 2.12 PF per trajectory, compared with 1.82 PF for the Qwen3 base. This estimate concerns encoder reuse; the reported throughput was measured with the current serving implementation.

\section{Per-Seed Results and Recency-Window Sweeps}
\label{app:seeds}

\subsection{Sampling Repeats}

Table~\ref{tab:seeds} lists the individual runs for both model families. Means and standard deviations are computed from the main-evaluation seeds. The additional-run column provides reference results for the recency-window sweeps below.

\begin{table}[tbp]
\centering
\caption{Per-seed resolved counts for Tables~\ref{tab:rq1} and~\ref{tab:transfer}, with additional reference runs. Counts are out of 499 unless noted; means and standard deviations use the main-evaluation seeds.}
\label{tab:seeds}
\footnotesize
\setlength{\tabcolsep}{4pt}
\begin{tabular}{@{}l r l r@{}}
\toprule
\textbf{Family / condition} & \textbf{additional run} & \textbf{main runs, per seed} & \textbf{mean $\pm$ sd} \\
\midrule
\multicolumn{4}{@{}l}{\textit{Qwen3-4B-Instruct-2507 family (Table~\ref{tab:rq1})}} \\
Uncompressed anchor & --- & 73 / 75 / 69 & 72.3 $\pm$ 3.1 (14.5\%) \\
LOHA plain text & 80 & 64 / 68 & 66.0 $\pm$ 2.8 (13.2\%) \\
LOHA Hard-Last-3 & 68 & 60 / 61 & 60.5 $\pm$ 0.7 (12.1\%) \\
Masking ($K{=}3$) & --- & 63 / 65 / 55 & 61.0 $\pm$ 5.3 (12.2\%) \\
\midrule
\multicolumn{4}{@{}l}{\textit{SWE-Master-4B-RL family (Table~\ref{tab:transfer})}} \\
Uncompressed teacher & 113 (of 494) & 133 / 141 & 137.0 $\pm$ 5.7 (27.5\%) \\
LOHA plain text & 128 & 117 / 126 & 121.5 $\pm$ 6.4 (24.3\%) \\
LOHA Hard-Last-3 & 111 & 114 / 104 & 109.0 $\pm$ 7.1 (21.8\%) \\
Masking ($K{=}3$) & --- & 88 / 87 / 89 & 88.0 $\pm$ 1.0 (17.6\%) \\
\bottomrule
\end{tabular}
\end{table}

\subsection{Recency-Window Sweeps}

The recency-window sweep in Table~\ref{tab:sweeps-main}a is detailed in Table~\ref{tab:sweeps}. Each condition uses one run, with one final result per instance.

\begin{table}[tbp]
\centering
\caption{Recency-window sweeps, one run per condition. Left: Qwen3-4B-Instruct-2507, 200-iteration cap, 262K context. Right: SWE-Master-4B-RL, 100-iteration cap, 131K context. Resolved counts are out of 499; dec PF is estimated decoder compute per trajectory.}
\label{tab:sweeps}
\footnotesize
\setlength{\tabcolsep}{5pt}
\begin{tabular}{@{}l rr rr@{}}
\toprule
& \multicolumn{2}{c}{\textbf{Qwen3-4B family}} & \multicolumn{2}{c}{\textbf{SWE-Master family}} \\
\cmidrule(lr){2-3}\cmidrule(l){4-5}
\textbf{Policy} & res. & dec PF & res. & dec PF \\
\midrule
plain text ($K{=}\infty$) & 80 & 1.47 & 128 & 2.99 \\
Hard-Last-8 & 72 & 2.10 & 115 & 8.37 \\
Hard-Last-4 & 76 & 2.08 & 108 & 4.49 \\
Hard-Last-3 & 68 & 1.67 & 111 & 3.74 \\
Hard-Last-2 & 65 & 1.40 & 103 & 2.93 \\
Hard-Last-1 & 47 & 1.21 & 88 & 1.89 \\
fully latent ($K{=}0$) & 37 & 1.18 & 51 & 0.93 \\
\bottomrule
\end{tabular}
\end{table}

For both models, resolve counts decrease substantially at $K{=}1$ and again at $K{=}0$. Increasing $K$ from 3 to 8 raises Qwen3's resolved count from 68 to 72 and SWE-Master's from 111 to 115, while estimated decoder compute rises from 1.67 to 2.10 PF and from 3.74 to 8.37 PF per trajectory. Larger windows retain more text and generally improve task performance, although the results are not monotonic in $K$: Qwen3 resolves 76 tasks at $K{=}4$. The diagnostic ablation in Appendix~\ref{app:results:rq3} separately compares $K{=}0$, $K{=}3$, and full text.

\section{Implementation and Training}
\label{app:method}

This section gives the rendering, parameterization, and optimization details for LOHA + ACD. The objectives are defined in Section~\ref{sec:acd}; their experimental analysis is in Appendix~\ref{app:results:rq2}.

\subsection{Context Rendering}

Both training views use the decoder's native chat template and tool-call format. System prompts, tool schemas, the issue, and assistant turns remain in text. Compressed tool observations keep their original positions and response delimiters; their bodies are marked as memory regions:

\begin{quote}
\ttfamily\raggedright
<|im\_start|>user\\
<tool\_response>\\
<|memory\_start|>\,\textrm{\itshape\ldots observation body\ldots}\,<|memory\_end|>\\
</tool\_response><|im\_end|>
\end{quote}

The 0.6B encoder processes each marked region in 1{,}024-token windows. The adapter maps it to $1/16$ as many soft tokens, while the surrounding tool envelope stays in text. No extra header is added: the preceding assistant tool call supplies the tool name and arguments.

In the latent training view, every observation body is encoded. In the full-text view, none is encoded. At inference, Hard-Last-$K$ determines which observations are compressed, with an additional rule retaining bodies shorter than 128 characters in text. The original trajectory's assistant tokens and the appended read-back answers are the supervised positions.

\subsection{Student Parameters}
\label{app:method:student}

The student combines the base decoder with the LCLM encoder and mean-pooling MLP adapter. The encoder is frozen; the 9.2M-parameter adapter is trainable. The reading patch adds rank-$\rho$ factors to every attention projection ($q,k,v,o$) and MLP projection (gate, up, and down): 252 matrices across 36 layers. Each factor pair is initialized from the rank-$\rho$ truncated singular value decomposition of $\Delta W=W_{\mathrm{LCLM}}-W_{\mathrm{base}}$.

At rank 64, the patch has 132M parameters. We also train 144 decoder normalization gain vectors, initialized at their base values, for a total of 141.5M trainable parameters. All other decoder parameters stay frozen. Teacher distributions come from the unmodified base model and are cached before training. Recovering that decoder from the student requires removing the patch and restoring the original normalization gains.

\subsection{Loss Aggregation}
\label{app:method:objective}

Let $\mathcal{A}_{\mathrm{L}}$ and $\mathcal{A}_{\mathrm{H}}$ be the supervised assistant tokens in latent and full-text examples, and $\mathcal{R}$ the read-back answer tokens. For answer token $j$, let $y_j$ be its target and $c_j^{\mathrm{R}}$ its preceding context. Let $\ell_{\mathrm{distill}}(j)$ and $\ell_{\mathrm{anchor}}(j)$ denote the per-token KL divergences in Equations~\ref{eq:acd-distill} and~\ref{eq:acd-anchor}, respectively. Then
\begin{align*}
\mathcal{L}_{\mathrm{distill}}
&= \frac{\displaystyle
\sum_{j\in\mathcal{A}_{\mathrm{L}}}\ell_{\mathrm{distill}}(j)
-\sum_{j\in\mathcal{R}}\log p_\theta(y_j\mid c_j^{\mathrm{R}})
}{|\mathcal{A}_{\mathrm{L}}|+|\mathcal{R}|},\\
\mathcal{L}_{\mathrm{anchor}}
&= \frac{1}{|\mathcal{A}_{\mathrm{H}}|}
\sum_{j\in\mathcal{A}_{\mathrm{H}}}\ell_{\mathrm{anchor}}(j).
\end{align*}
The total loss is $\mathcal{L}_{\mathrm{distill}}+\lambda\mathcal{L}_{\mathrm{anchor}}$, with $\lambda{=}1$ in all reported runs. Each term is averaged over its own supervised tokens. The latent loss updates the adapter, patch, and gains; the full-text anchor updates only the patch and gains. Targets are aligned by assistant-token identity within a trajectory, since absolute positions differ between the rendered views.

\subsection{Cached Teacher Distributions}

We score each full-text trajectory once and cache the teacher's top-$M$ token log-probabilities and residual probability mass at every assistant position. The default is $M{=}64$; the full corpus contains 0.31B scored positions and uses 56 GB of cached targets. Each record is keyed by trajectory and assistant span and joined to the training examples during packing.

Both KL terms use the cached tokens plus one tail bucket. For teacher probabilities $p_i$, student probabilities $q_i$, residual teacher mass $r=1-\sum_{i\le M}p_i$, and student tail mass $q_{\mathrm{tail}}=1-\sum_{i\le M}q_i$, the forward and reverse estimates are
\begin{align*}
D_{\rightarrow}
&=\sum_{i\le M}p_i\log\frac{p_i}{q_i}
+r\log\frac{r}{q_{\mathrm{tail}}},\\
D_{\leftarrow}
&=\sum_{i\le M}q_i\log\frac{q_i}{p_i}
+q_{\mathrm{tail}}\log\frac{q_{\mathrm{tail}}}{\max(r,\rho_{\min})}.
\end{align*}
Before numerical stabilization, coarsening gives lower bounds on the full-vocabulary divergences by the log-sum inequality. Equality holds when teacher and student have the same relative probabilities within the tail.

The early reference ablation uses top-16 targets, as marked in Table~\ref{tab:rq2b}. Reverse-KL variants use $\rho_{\min}{=}10^{-6}$ and clip per-token log-ratios at 10. The default recipe uses forward KL for both objectives. Approximation accuracy is measured in Appendix~\ref{app:mechanism:kl}.

\subsection{Read-back Supervision}

Latent-view trajectories include read-back questions at a density of one question per four observations. Questions are generated by tool type and ask for literal content such as a viewed code range, matching search lines, a failing test and assertion, a traceback frame, or an exit code. Candidate edit strings can also come from the trajectory's \texttt{old\_str} arguments.

Candidates are filtered so that the answer is absent from hard-token context and cannot be inferred from the question text alone. Answer tokens receive next-token cross-entropy and account for 3\% of supervised positions. The original trajectory tokens receive the distributional objectives above.

\subsection{Data and Optimization}

We select the shortest successful trajectory per task from CoderForge~\citep{together2026coderforge}, giving 35{,}140 trajectories, 1.29B raw tokens, and 0.31B teacher-scored assistant positions. The Qwen3 diagnostic, teacher, and objective ablations use the first 3{,}514 trajectories under a fixed ordering. We apply no additional repository filter beyond the corpus's construction.

Both views are packed into 204{,}800-token sequences with span-level masks identifying distillation, anchor, and read-back supervision. The build checks that full-text rendering reproduces the teacher-scored token sequence; trajectories failing this check are dropped. Teacher arrays are checked against the student positions at load time.

We train for one epoch using AdamW with $\beta=(0.9,0.95)$, a cosine schedule, and 10\% warm-up. Learning rates are $10^{-4}$ for patch factors and normalization gains, and $5{\times}10^{-5}$ for the adapter. On the ablation corpus, one epoch consists of 125 optimizer steps of approximately 1.6M decoder tokens on eight GPUs. Every trajectory appears in both views during the epoch, and cached targets avoid a teacher forward pass during optimization.

\subsection{Behavioral Probes}
\label{app:method:gates}

Candidate models are evaluated on the 96-field recitation and 48-field structured extraction probes, followed by a plain-text tool-calling probe on real tasks. The latter compares edits per trajectory and well-formed tool calls with the base model. It serves as a release criterion, but all reported ablation arms were evaluated regardless of their probe results. This separates reading quality from the ability to act with the scaffold's tools.

\subsection{Inference and Caching}
\label{app:method:inference}

The agent uses the existing tool schemas and native function-calling harness. Before each call, LOHA keeps the last $K$ observations and short bodies in text and marks older bodies for encoding. The server encodes those regions when the request arrives. The current implementation re-encodes them on every call; Appendix~\ref{app:results:rq4} evaluates the savings from reusing encoder outputs.

Fully latent histories are append-only. Under Hard-Last-$K$, an observation changes from text to soft tokens when it leaves the window. This invalidates decoder prefix-cache entries from the first changed observation onward, so the recent tail and intervening assistant turns must be recomputed. The encoder and decoder cache effects are measured separately.

\section{Diagnostic Measurements}
\label{app:mechanism}

These measurements characterize the released LCLM decoder, its latent inputs, and the reading patch. They complement the agent-level ablations with direct tests of representations and recall.

\begin{figure}[t]
\centering
\includegraphics[width=\textwidth]{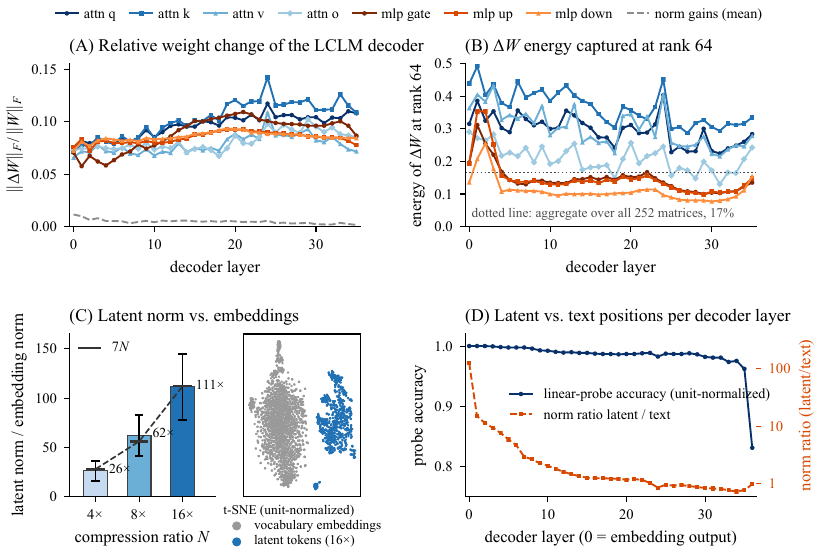}
\caption{Measurements of the LCLM decoder and latent representations. (A)~Relative Frobenius change of each projection matrix from the base to the LCLM decoder. (B)~Fraction of the energy of $\Delta W$ in its rank-64 truncation. (C)~Norm of the adapter's soft tokens relative to the word embeddings at three compression ratios, and a t-SNE of latent and vocabulary embeddings. (D)~Per-layer accuracy of a probe separating latent from text positions, and the ratio of their norms.}
\label{fig:mechanism-full}
\end{figure}

\subsection{Measurement Setup}

Figure~\ref{fig:mechanism-full} compares the released $16\times$ LCLM decoder with Qwen3-4B-Instruct-2507. Encoder--adapter measurements use 48 windows of 1{,}024 tokens from real tool observations at compression ratios $4\times$, $8\times$, and $16\times$.

Panels A and B examine the 252 projection matrices. For $\Delta W=W_{\mathrm{LCLM}}-W_{\mathrm{base}}$, panel A plots $\|\Delta W\|_F/\|W\|_F$ and the mean change in four normalization gains per layer. Panel B reports the fraction of $\|\Delta W\|_F^2$ retained by rank-64 truncation; the aggregate is weighted by each matrix's squared change norm.

Panel C measures latent-token norms relative to mean vocabulary-embedding norm, with one-standard-deviation error bars. Its t-SNE uses 2{,}000 vocabulary embeddings and 1{,}000 latent embeddings after unit normalization, PCA to 50 dimensions, and perplexity 30. Panel D uses five-fold cross-validated logistic regression to distinguish 512 latent positions from 1{,}152 text positions in unit-normalized residual-stream vectors at the embedding output and all 36 decoder layers.

\subsection{Weight Changes and Reading Initialization}

Projection matrices change by 6--14\% of their Frobenius norm across layers. Normalization gains change by about 1\% or less outside the first layer. The effective rank of $\Delta W$ ranges from 0.69 to 0.94 of full rank; rank-64 truncation retains 17\% of total change energy, with a per-matrix median of 22\%.

Applied to the base decoder, the rank-64 patch recovers 79\% of the LCLM decoder's structured read-back score, while exact literal recitation reaches 27\% of its rate. The same untrained patch degrades agent behavior. Thus the initialization recovers substantial reading ability, but its effect on acting still needs to be addressed during adaptation.

\subsection{Latent and Text Representations}

Mean adapter-output norms are $26\times$, $62\times$, and $111\times$ the word-embedding norm at $4\times$, $8\times$, and $16\times$ compression. Unit-normalized latent and vocabulary embeddings occupy separate regions in the t-SNE visualization. The linear probe distinguishes latent from text positions with 98--100\% accuracy through layer 35, even after their norms have converged. These measurements show that latent and text positions remain distinguishable throughout most of the decoder.

\subsection{Gradient and Teacher Diagnostics}

At the warm-started patch, gradients of latent-view read-back and full-text next-token objectives have correlations of 0.93 and 0.95 in two measurements. A mask selecting directions with reading gain and low full-text cost retains 2{,}559 of 16{,}128 directions across the 252 rank-64 factor pairs. It recovers eight of 48 structured fields, against seven for the unpatched base and 22 for the full patch. The tested selection therefore removes most of the patch's reading gain.

Disagreement between the 30B teacher and 4B base model concentrates on 15.5\% of positions with high-entropy next-action predictions. Their top-1 agreement is 39--65\% at those positions and nearly complete elsewhere. The corresponding agent-level comparison appears in Appendix~\ref{app:results:rq2}, where behavioral anchoring and the distillation teacher are varied separately.

\subsection{Cached-Distribution Approximation}
\label{app:mechanism:kl}

On held-out positions at $M{=}32$, the cached forward-KL estimate is within 0.1\% of exact KL. The reverse estimate underestimates by 3.5--4.2\%, with gradient cosine 0.99; its gap concentrates on positions with student mass outside the cached tokens and approximately halves as $M$ doubles.

\subsection{Exact Recitation}
\label{app:recitation}

We sample 358 literals that later actions quote in real trajectories, including \texttt{old\_str} arguments, \texttt{pytest} identifiers, search patterns, and embedded code. The LCLM decoder receives the latent context and is scored on exact recitation, using the same sample at every compression ratio.

Accuracy is 20.4\% at $4\times$, 15.6\% at $8\times$, and 15.1\% at $16\times$. No literal-type and length bucket with at least 10 examples reaches 90\%. Composed \texttt{pytest} identifiers are never exact, and literals of at least 96 characters are almost never exact. Near-misses such as \texttt{description} rendered as \texttt{descript} illustrate why semantic access to an observation does not ensure exact tool arguments.

\section{Extended Related Work}
\label{app:related}

\subsection{Software-Engineering Agents}

SWE-bench provides executable repair tasks from real GitHub issues~\citep{jimenez2024swebench}. SWE-agent~\citep{yang2024sweagent}, OpenHands~\citep{wang2024openhands}, AutoCodeRover~\citep{zhang2024autocoderover}, and Agentless~\citep{xia2024agentless} combine repository exploration, editing, and testing. CoderForge~\citep{together2026coderforge} and SWE-smith~\citep{yang2025swesmith} provide trajectories for training coding agents. Our work retains the tool interaction pattern and changes the representation of accumulated observations and the training needed to use it.

\subsection{Context Management and Pruning}

Observation masking can substantially reduce cost while retaining performance comparable to history summarization~\citep{lindenbauer2025complexity}. Other approaches learn compression guidelines~\citep{kang2025acon}, fold completed sub-trajectories~\citep{sun2025contextfolding}, restructure the workspace~\citep{ye2025agentfold}, page history through a hierarchy~\citep{packer2023memgpt}, or organize it as a dependency graph~\citep{wu2026contextweaver}.

Methods targeting coding observations include SWE-Pruner's goal-conditioned line selection~\citep{wang2026swepruner}, SWE-Pruner Pro's pruning head over coder hidden states~\citep{wang2026sweprunerpro}, TACO's evolved filtering rules~\citep{ren2026taco}, and CoACT's action-preserving compression targets~\citep{chen2026coact}. LOHA instead represents older observation bodies with soft tokens and keeps a recent text window. Masking supplies a direct comparison because it uses the same recency structure while removing the older content.

\subsection{Prompt and Code Compression}

Token-space compressors remove low-information tokens or spans~\citep{li2023selectivecontext,jiang2023llmlingua,pan2024llmlingua2}. LongCodeZip uses function-level structure and conditional perplexity~\citep{shi2025longcodezip}, while SlimCode simplifies code inputs independently of the downstream model~\citep{wang2024slimcode}. In an agent trajectory, a later edit or command may require an identifier or code fragment that was not salient when the observation arrived. LOHA separates recent exact text from a compressed history; the comparison with LongCodeZip examines this alternative on the same observation stream.

\subsection{Latent Compression and Memory}

Gist tokens~\citep{mu2023gist}, in-context autoencoders~\citep{ge2024icae}, and extreme compressors~\citep{li2024500xcompressor} establish the interface between compressed embeddings and a language-model decoder. LCLM scales this approach with a 0.6B encoder and 4B decoder trained on 350B tokens~\citep{li2026lclm}. We use that substrate for a multi-turn agent whose observations include code, search results, and test output.

Latent memory systems also encode GUI trajectories~\citep{zhang2026memw}, generate memory during reasoning~\citep{zhang2025memgen}, or specialize it by agent role~\citep{fu2026latentmem}. These address experience reuse and coordination. LOHA focuses on observations within the current trajectory, including the exact text needed for coding-tool arguments.

\subsection{Distillation and Behavioral Anchoring}

Knowledge distillation transfers model predictions to a student~\citep{hinton2015distilling,kim2016sequence}. Context distillation uses a richer teacher context to supervise a reduced student context~\citep{askell2021general,snell2022context}; xRAG applies this idea to compressed retrieved documents~\citep{cheng2024xrag}. ACD applies cross-view distillation at corresponding assistant tokens in tool-use trajectories. Work on teacher--student capacity gaps~\citep{mirzadeh2020improved}, imitation of stronger models~\citep{gudibande2024false}, and same-capacity teachers~\citep{furlanello2018born} motivates examining the reference model explicitly. Reverse-KL and on-policy distillation offer alternative objectives~\citep{gu2024minillm,agarwal2024gkd}.

The full-text anchor is related to Learning without Forgetting~\citep{li2018lwf}, self-distillation fine-tuning~\citep{yang2024sdft}, and KL regularization toward a reference policy~\citep{ziegler2019finetuning,ouyang2022training}. Weight-space methods such as elastic weight consolidation instead constrain parameter changes~\citep{kirkpatrick2017ewc}. ACD combines a distributional behavior anchor with latent-view distillation and trains a low-rank patch~\citep{hu2022lora} initialized from the compressor's weight difference. The two objectives have distinct roles: one teaches the compressed representation, while the other regularizes the adapted agent toward its own base policy.

\end{document}